# A Case Study in Complexity Estimation: Towards Parallel Branch-and-Bound over Graphical Models

**Lars Otten** and **Rina Dechter**
Department of Computer Science
University of California, Irvine
{lotten,dechter}@ics.uci.edu

## Abstract

We study the problem of complexity estimation in the context of parallelizing an advanced Branch and Bound-type algorithm over graphical models. The algorithm's pruning power makes load balancing, one crucial element of every distributed system, very challenging. We propose using a statistical regression model to identify and tackle disproportionally complex parallel subproblems, the cause of load imbalance, ahead of time. The proposed model is evaluated and analyzed on various levels and shown to yield robust predictions. We then demonstrate its effectiveness for load balancing in practice.

## 1 INTRODUCTION

This paper explores the application of learning for improved load balancing in the context of distributed search for discrete combinatorial optimization over graphical models (e.g., Bayesian networks, weighted CSPs). Specifically, we consider one of the best exact search algorithms for solving the MPE task over graphical models, AND/OR Branch and Bound (AOBB) [12], ranked first and third, respectively, in the UAI'06 and '08 evaluations and winning all three MPE categories of the 2011 PASCAL Inference Challenge.

We adapt the established concept of parallel tree search [7], where a search tree is explored centrally up to a certain depth and the remaining subtrees are solved in parallel. In the graphical model context we explore the search space of partial instantiations up to a certain point and solve the resulting conditioned subproblems in parallel.

The distributed framework is built with a grid computing environment in mind, i.e., a set of autonomous, loosely connected systems – notably, we cannot assume any kind of shared memory or dynamic load balancing which most parallel or distributed algorithms build upon [1, 5, 7, 6]. The primary challenge is therefore to determine a priori a set of subproblems with balanced complexity, so that the overall parallel runtime will not be dominated by just a few of them. In the optimization context, however, the use of cost and heuristic functions for pruning makes it very hard to reliably predict and balance subproblem complexity ahead of time; in particular, structural parameters like the induced width are not sufficient to differentiate subproblems.

Our suggested approach and the main contribution of this paper is to estimate subproblem complexity by learning a regression model over several subproblem parameters, some static and structural (e.g., induced width, variable domain sizes), others dynamically extracted at runtime (e.g. upper and lower bounds on the subproblem solution based on the heuristic function).

A similar regression-based approach was developed in [11] to predict the problem complexity (called "empirical hardness") of combinatorial auction instances; similarly the successful SAT solver *SATzilla* uses linear regression models to choose among a set of component solvers the one that is predicted to be fastest for a given SAT instance [16].

Other general work on estimating search complexity goes back to [10] and more recently [9], which predict the size of general backtrack trees through random probing. Similar schemes were devised for Branch and Bound algorithms [2], where search is run for a limited time and the partially explored tree is extrapolated. These approaches typically require a substantial amount of probing, which is prohibitively expensive in our setup, where many hundreds, if not thousands of subproblems need to be evaluated quickly.

The contribution of the present paper lies in proposing and studying a general learning approach for estimating subproblem complexity. In particular, we frame the problem as statistical regression analysis, which allows us to leverage established, powerful techniques from machine learning and statistics. Motivated by different parallelization scenarios, we distinguish three distinct levels of learning: based on a single problem instance, based on a specific class of problems, and based on a combination of problem classes. We evaluate, analyze, and contrast these three levels on a sample set of more than 11,000 subproblem samples from four problem classes and demonstrate generally robust prediction performance. We also demonstrate empirically the model's potential for improved load balancing.

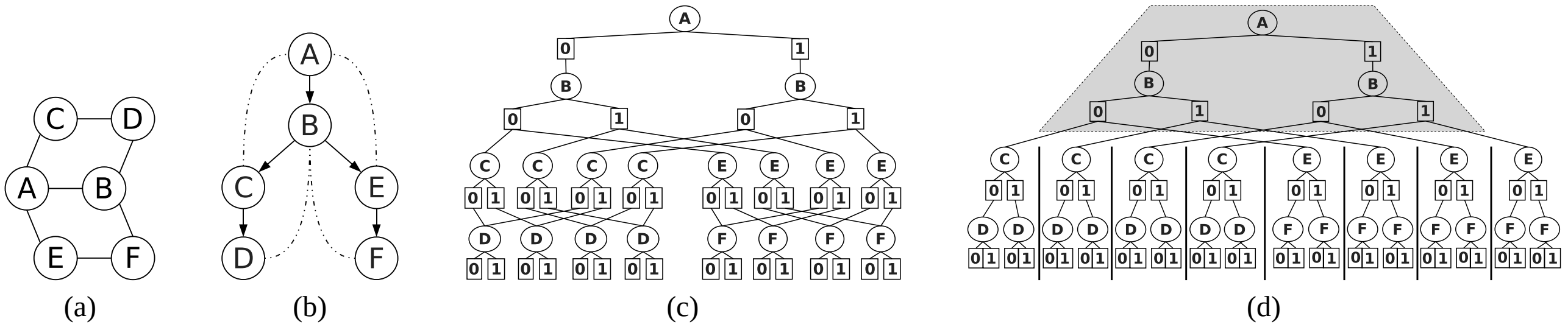

Figure 1: (a) Example primal graph with six variables, (b) its pseudo tree along ordering $A, B, C, D, E, F$, (c) the corresponding context-minimal AND/OR search graph, and (d) the parallel search space resulting from parallelizing at depth $d = 2$ with eight independent subproblems.

The remainder of the paper is organized as follows: Section 2 summarizes the necessary background and outlines the distributed AND/OR Branch and Bound algorithm. Section 3 introduces the proposed regression model for complexity estimation while Section 4 evaluates it on a variety of instances from several problem classes. Section 5 provides selected parallel results that highlight the benefits of the proposed model and Section 6 concludes.

## 2 BACKGROUND

We assume the usual definitions of a *graphical model* as a set of functions $F = \{f_1, \ldots, f_m\}$ over discrete variables $X = \{X_1, \ldots, X_n\}$, its *primal graph*, *induced graph*, and *induced width*.

### 2.1 AND/OR SEARCH SPACES

The concept of AND/OR search spaces has been introduced as a unifying framework for advanced algorithmic schemes for graphical models to better capture the structure of the underlying graph [3]. Its main virtue consists in exploiting conditional independencies between variables, which can lead to exponential speedups. The search space is defined using a *pseudo tree*, which captures problem decomposition:

DEFINITION **1 (pseudo tree)** *Given an undirected graph $G = (X, E)$, a* pseudo tree *of $G$ is a directed, rooted tree $\mathcal{T} = (X, E')$ with the same set of nodes $X$, such that every arc of $G$ that is not included in $E'$ is a back-arc in $\mathcal{T}$, namely it connects a node in $\mathcal{T}$ to an ancestor in $\mathcal{T}$. The arcs in $E'$ may not all be included in $E$.*

**AND/OR Search Trees and Graphs :** Given a graphical model instance with variables $X$ and functions $F$, its primal graph $(X, E)$, and a pseudo tree $\mathcal{T}$, the associated *AND/OR search tree* consists of alternating levels of OR and AND nodes. The structure of the AND/OR search tree is based on the underlying pseudo tree $\mathcal{T}$: the root of the AND/OR search tree is an OR node labeled with the root of $\mathcal{T}$. The children of an OR node $X_i$ are AND nodes labeled with assignments $\langle X_i, x_i \rangle$; the children of an AND node $\langle X_i, x_i \rangle$ are OR nodes labeled with the children of $X_i$ in $\mathcal{T}$, representing conditionally independent subproblems. Different nodes may root identical subproblems and can be merged through *caching*, yielding an *AND/OR search graph* of smaller size, at the expense of using additional memory during search.

Given a graphical model, its primal graph $G$, and a guiding pseudo tree $\mathcal{T}$ of height $h$, the size of the AND/OR search tree is $\mathcal{O}(n \cdot k^h)$, while $\mathcal{O}(n \cdot k^{w^*})$ bounds the AND/OR search graph, where $w^*$ is the induced width of $G$ over a depth-first traversal of $\mathcal{T}$ and $k$ bounds the domain size [3]. Figure 1(a) shows an example problem primal graph with six variables, Figure 1(b) depicts a pseudo tree along ordering $A, B, C, D, E, F$. Figure 1(c) shows the corresponding AND/OR search graph.

**AND/OR Branch and Bound :** AND/OR Branch and Bound (AOBB) is a state-of-the-art algorithm for solving optimization problems over graphical models. Assuming maximization, it traverses the AND/OR graph in a depth-first manner while keeping track of a current lower bound on the optimal solution cost. During expansion of a node $n$, this lower bound $l$ is compared with a heuristic upper bound $u(n)$ on the optimal solution below $n$ – if $u(n) \leq l$ the algorithm can prune the subproblem below $n$ [12].

**Mini-Bucket Heuristics :** The heuristic $h(n)$ that we use in our experiments is the Mini-Bucket heuristic. It is based on Mini-Bucket elimination, an approximate variant of a variable elimination scheme that computes approximations to reasoning problems over graphical models [4]. A control parameter $i$ allows to trade accuracy of the heuristic against its time and space requirements. It was shown that the intermediate functions generated by the Mini-Bucket algorithm MBE($i$) can be used to derive a heuristic function that un-derestimates the minimal cost solution to a subproblem in the AND/OR search graph [12].

### 2.2 DISTRIBUTED AOBB & LOAD BALANCING

Our distributed implementation of AND/OR Branch and Bound draws from the notion of parallel tree search [7, 6], where a search tree is explored centrally up to a certain depth and the remaining subtrees are solved in parallel. Applied to the search graph from Figure 1(c), for instance, we could obtain eight independent subproblems as shown in Figure 1(d), with a conditioning search space (in gray) spanning the first two levels (variables $A$ and $B$).

We refer to the boundary between conditioning search space and parallel subproblems as the *parallelization frontier*. Its choice determines the shape and the number of subproblems and is thus crucial for effective parallel *load balancing*. Namely, it is known that for best parallel performance we should spread the parallel workload evenly across all available CPUs, while minimizing overhead. Note that we assume independent worker machines, with limited or very costly communication, hence dynamic load balancing at runtime (cf. [6]) is not applicable.

Algorithm 1 shows pseudo code for our parallelization policy: the parallelization frontier is generated in a breadth-first manner by iteratively selecting the current most complex subproblem, estimated by a complexity estimator $\hat{N}$, and splitting it into its immediate "sub-subproblems", which are in turn added to the frontier. This process is repeated until a desired number of subproblems is obtained, at which point all subproblems are submitted to the distributed environment.

In the context of depth-first Branch and Bound, however, determining the most complex subproblem is extremely difficult and elusive. Due to the pruning power of the algorithm, subproblem runtimes can differ greatly, even when the underlying subgraph structure and the associated asymptotic complexity guarantees (exponential in the induced width of the AND/OR subspace) are identical.

To illustrate, consider the subproblem statistics of two parallel runs shown in Figure 2, where instead the parallelization frontier is placed at a fixed depth $d = 5$ and $d = 4$, respectively, yielding 64 and 144 subproblems (the horizontal axis). In each case we see significant variance in subproblem runtime. In fact, the overall runtime is dominated exclusively by the handful of longest-running subproblems, with most other subproblems finishing long before (note the log scale). Detecting and mitigating this imbalance ahead of time constitutes the central challenge in this line of work, as we elaborate in the next sections.

**Algorithm 1** Finding the parallelization frontier

**Input:** Pseudo tree $\mathcal{T}$ with root $X_0$, subproblem count $p$, subproblem complexity estimator $\hat{N}$ .

**Output:** Set $F$ of subproblem root nodes with $|F| \geq p$ .

$F \leftarrow \{\langle X_0 \rangle\}$

**while** $|F| < p$ :

  $n' \leftarrow \arg\max_{n \in F} \hat{N}(n)$

  $F \leftarrow F \setminus \{n'\}$

  $F \leftarrow F \cup children(n')$

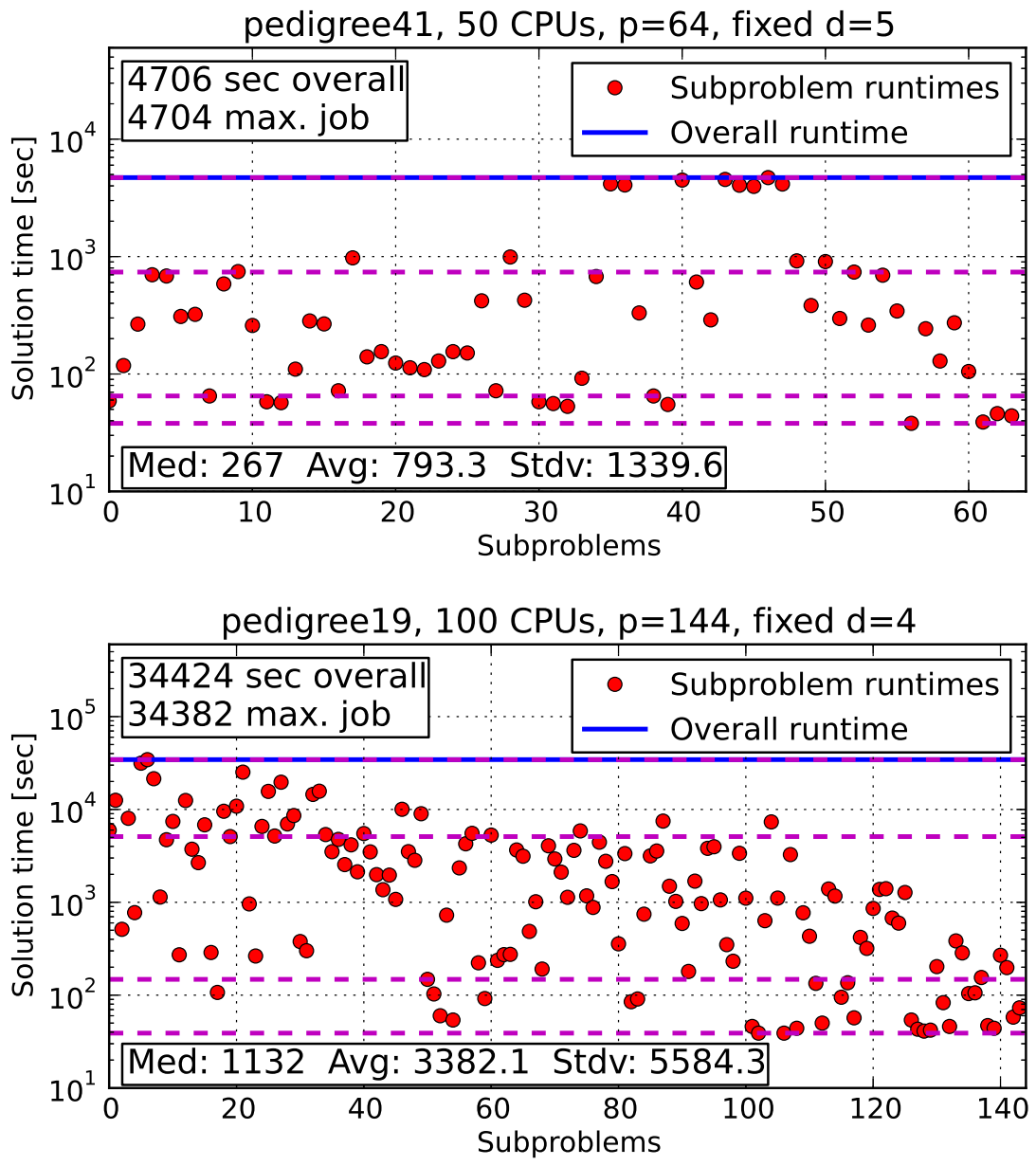

Figure 2: Subproblem statistics for fixed-depth parallelization frontier showing large variance in subproblem runtime. Dashed lines mark 0, 20, 80, and 100 percentile.

## 3 LEARNING COMPLEXITIES THROUGH REGRESSION

This section introduces our learning approach to subproblem complexity prediction through regression analysis. Previous work has investigated and evaluated various methods for balancing subproblem complexity, directly formulating metrics using human expert knowledge [13, 14]. These metrics were relative in nature, i.e., they only allowed comparison of one subproblem to another within a given overall problem instance. In contrast, the present work does not depend as heavily on expert knowledge and gives absolute complexity estimates.

### 3.1 GENERAL METHODOLOGY

We identify a subproblem by its search space root node $n$. We further measure the complexity of the subproblem rooted at $n$ through the size of its explored search space, which is the number of node expansions required for its solution, denoted $N(n)$. We then aim to capture the exponential nature of the search space size by modeling $N(n)$ as an exponential function of various subproblem features $\phi_i(n)$ as follows:

$$N(n) = \exp(\sum_i \lambda_i \phi_i(n)) \tag{1}$$

The exponent has been chosen as a sum so that we can consider the log complexity and obtain the following:

$$\log N(n) = \sum_i \lambda_i \phi_i(n) \tag{2}$$

Given a set of $m$ sample subproblems, finding suitable parameter values $\lambda_j$ can thus be formulated as a well-known *linear regression* problem, with the *mean squared error* (MSE) as the loss function $L(\lambda)$ we aim to minimize:

$$L(\lambda) = \frac{1}{m} \sum_{k=1}^{m} \Big( \sum_i \lambda_i \phi_i(n_k) - \log N(n_k) \Big)^2 \tag{3}$$

The MSE captures how well the learned regression model fits the training data. In the context of load balancing for parallelism we can consider a secondary metric, the *Pearson correlation coefficient* (PCC), which is simply the normalized covariance between the vector of subproblem complexities and their estimates, normalized by the product of each vector's standard deviation. It is bounded by $[-1, 1]$, where 1 implies perfect linear correlation and -1 anticorrelation. Hence a value close to 1 is desirable, as it signifies a model likely to correctly identify the hardest subproblems.

### 3.2 SUBPROBLEM FEATURES

Table 1 lists the full set of basic subproblem features $\phi_i$ that we consider. This list was compiled based on our prior knowledge of what aspects can affect problem complexity. Features can be divided into two distinct classes: "static", which can be precompiled from the problem graph and pseudo tree, and "dynamic" which are computed at runtime, as the parallelization frontier decision is made (note that none of the dynamic features are costly to compute).

| | |
|---|---|
| **Subproblem variable statistics (static):** | |
| 1: | Number of variables in subproblem. |
| 2-6: | Min, Max, mean, average, and std. dev. of variable domain sizes in subproblem. |
| **Pseudotree depth/leaf statistics (static):** | |
| 7: | Depth of subproblem root in overall search space. |
| 8-12: | Min, max, mean, average, and std. dev. of depth of subproblem pseudo tree leaf nodes, counted from subproblem root. |
| 13: | Number of leaf nodes in subproblem pseudo tree. |
| **Pseudo tree width statistics (static):** | |
| 14-18: | Min, max, mean, average, and std. dev. of induced width of variables within subproblem. |
| 19-23: | Min, max, mean, average, and std. dev. of induced width of variables within subproblem, *when conditioning on subproblem root conditioning set*. |
| **Subproblem cost bounds (dynamic):** | |
| 24: | Lower bound $L$ on subproblem solution cost, derived from current best overall solution. |
| 25: | Upper bound $U$ on subproblem solution cost, provided by mini bucket heuristics. |
| 26: | Difference $U - L$ between upper and lower bound, expressing "constrainedness" of the subproblem. |
| **Pruning ratios (dynamic)**, based on running 5000 node expansion probe of AOBB: | |
| 27: | Ratio of nodes pruned using the heuristic. |
| 28: | Ratio of nodes pruned due of determinism (zero probabilities, e.g.) |
| 29: | Ratio of nodes corresponding to pseudo tree leaf. |
| **Sample statistics (dynamic)**, based on running 5000 node expansion probe of AOBB: | |
| 30: | Average depth of terminal search nodes within probe. |
| 31: | Average node depth within probe (denoted $\bar{d}$). |
| 32: | Average branching degree, defined as $\sqrt[\bar{d}]{5000}$. |
| **Various (static):** | |
| 33: | Mini bucket $i$-bound parameter. |
| 34: | Max. subproblem variable context size minus mini bucket $i$-bound. |

Table 1: Subproblem features for complexity estimation.

### 3.3 SUBPROBLEM SAMPLE DOMAINS

In order to evaluate training and prediction error of the proposed complexity model from a statistical learning perspective, we need to specify the sample domain over which we will make predictions, for which we aim to generate a model, and from which subproblem samples are assumed to be drawn. In fact, in the following we consider three incrementally more general levels of sample domains and learning, corresponding to three different designs in the context of parallelizing AOBB:

1. **Learning per problem instance:** The sample domain is all subproblems from a single problem instance. This corresponds to learning a new complexity model for every problem instance the parallel scheme encounters (e.g., we would learn separate models for the two instances in Figure 2).

2. **Learning per problem class:** Take the domain to be all subproblems of problems from a specific class. In the parallelization context we learn a separate model for every problem class we consider. For example, we would learn a single model for all pedigree problems, but a different model for other problem classes like protein sidechain prediction.

3. **Learning across problem classes:** Take the sample domain to be all subproblems of all problems from several classes. Ultimately this could translate to a parallel scheme that uses a single complexity model for all problem classes under consideration.

These three levels are increasingly more general and thus potentially more challenging for robust estimation. On the other hand, they require increasingly less computational ef-fort, since fewer distinct models need to be learned. Lastly, they can present different trade-offs between pre-compiled off-line learning and learning at runtime.

### 3.4 REGRESSION ALGORITHMS

We investigated a number of algorithms for fitting a linear model. Ordinary least squares (OLS) regression was problematic due to numerical issues (near-singular matrix inversion) and prone to overfitting (due to lack of regularization) and we did not consider if further. Standard ridge regression adds the $L_2$-norm of the parameter vector $\lambda$ to the regularized loss function through a term $\alpha(\sum_i \lambda_i^2)^{\frac{1}{2}}$; similarly, lasso regression [15], places an $L_1$-penalty on the parameter vector by adding the term $\alpha \sum_i |\lambda_i|$ . The so-called "Elastic Net" combines both penalty terms [18]. In each case we followed the common approach of determining the regularization parameter $\alpha$ once through initial cross validation and held it fixed subsequently.

In our experiments we found all methods to perform similarly in terms of training and prediction errors, with a slight advantage for the lasso method. We will therefore focus on lasso learning. This method has the additional benefit of "built-in" feature selection: learned models are relatively sparse and thus compact, because the $L_1$-regularization pushes many parameters $\lambda_i$ to zero [15].

### 3.5 NON-LINEAR REGRESSION

In addition to the purely linear regression analysis proposed above, we also explored non-linear approaches. In particular, we took inspiration from [11], which reports improved prediction performance using *quadratic feature expansion*, albeit in the context of combinatorial auctions. Quadratic feature expansion, also referred to as "quadratic regression", works by adding new features in the form of pairwise products of the original features; namely, for every pair of subproblem features $\phi_i$, $\phi_j$ with $i \leq j$, we create a new feature $\phi_i \cdot \phi_j$. We then perform linear regression on the expanded feature set (629 in our case), thereby effectively fitting a polynomial of 2nd degree. Results will be outlined in Section 4.5.

Next we evaluate the proposed regression model on a variety of instances from several different problem classes.

## 4 EVALUATION AND ANALYSIS

The basis for our evaluation are 31 hard problem instances from four classes: pedigree haplotyping problems, protein side-chain prediction ([17], named "pdb"), "large family" genetic linkage instances, and grid networks ("75-2x-x"). Summary statistics of the different problem classes are given in Table 2. We note that all instances each take several hours, if not days to solve using sequential AOBB.

| domain | $M$ | $n$ | $k$ | $w$ | $h$ |
|---|---|---|---|---|---|
| pedigree | 13 | 437 – 1272 | 3 – 7 | 17 – 39 | 47 – 102 |
| pdb | 5 | 103 – 172 | 81 | 10 – 15 | 24 – 43 |
| largeFam | 8 | 2569 – 3730 | 3 – 4 | 28 – 37 | 73 – 108 |
| grid | 5 | 624 – 675 | 2 | 37 – 39 | 111 – 124 |

Table 2: Summary statistics for problem classes used, $M$ gives the number of instances in the class. $n$ denotes number of problem variables, $k$ max. domain size, $w$ induced width, $h$ pseudo tree height.

To compile a set of subproblem samples we revisit experiments with fixed-depth parallelization (cf. Section 2.2): we randomly choose not more than 500 subproblems from a previously recorded fixed-depth parallel run for each instance. This leaves us with about 11,500 sample subproblems (approx. 40% pedigree, 25% protein, 25% largeFam, 10% grids), which is very reasonable for the number of features we have (the variance of the trained linear model scales with $p/m$, where $p$ is the number of features and $m$ the number of samples, cf. Section 7.3 in [8]).

The empirical evaluation is organized as follows: Sections 4.1 through 4.3 assess the prediction power of our proposed linear regression complexity model according to the three levels of learning outlined in Section 3.3. Section 4.4 inspects feature informativeness and Section 4.5 briefly investigates performance of the quadratic model. Section 4.6 provides a summary of the learning results.

Throughout this section results are presented as a log-log scatter plot of actual versus predicted complexities, each also containing mean squared prediction error ("MSE", on the test set) and Pearson correlation coefficient ("PCC") as well as mean squared training error ("TER").

### 4.1 LEARNING PER PROBLEM INSTANCE

This first set of experiments is meant to determine the prediction quality of a regression model that is learned for a single instance only. To that end, we consider all subproblem samples from a given problem instance and apply 5-fold cross validation (i.e., partition the samples into 5 subsets, then predict the complexities of each subset by learning a model on the remaining four).

Figure 3 presents scatter plots for six problem instances from the different problem classes considered. We see that results are good for the protein and largeFam instance and still acceptable for *pedigree19* with slightly higher MSE. *Pedigree41* has a relatively low MSE and good PCC, in spite of the plot's flat appearance. In case of the grid instance *75-26-9* the model's discriminatory power is likely limited by the small number of subproblem samples in this case. Finally, we note that the training error ("TER") is very close to the prediction error ("MSE") in all cases, indicating the absence of overfitting.

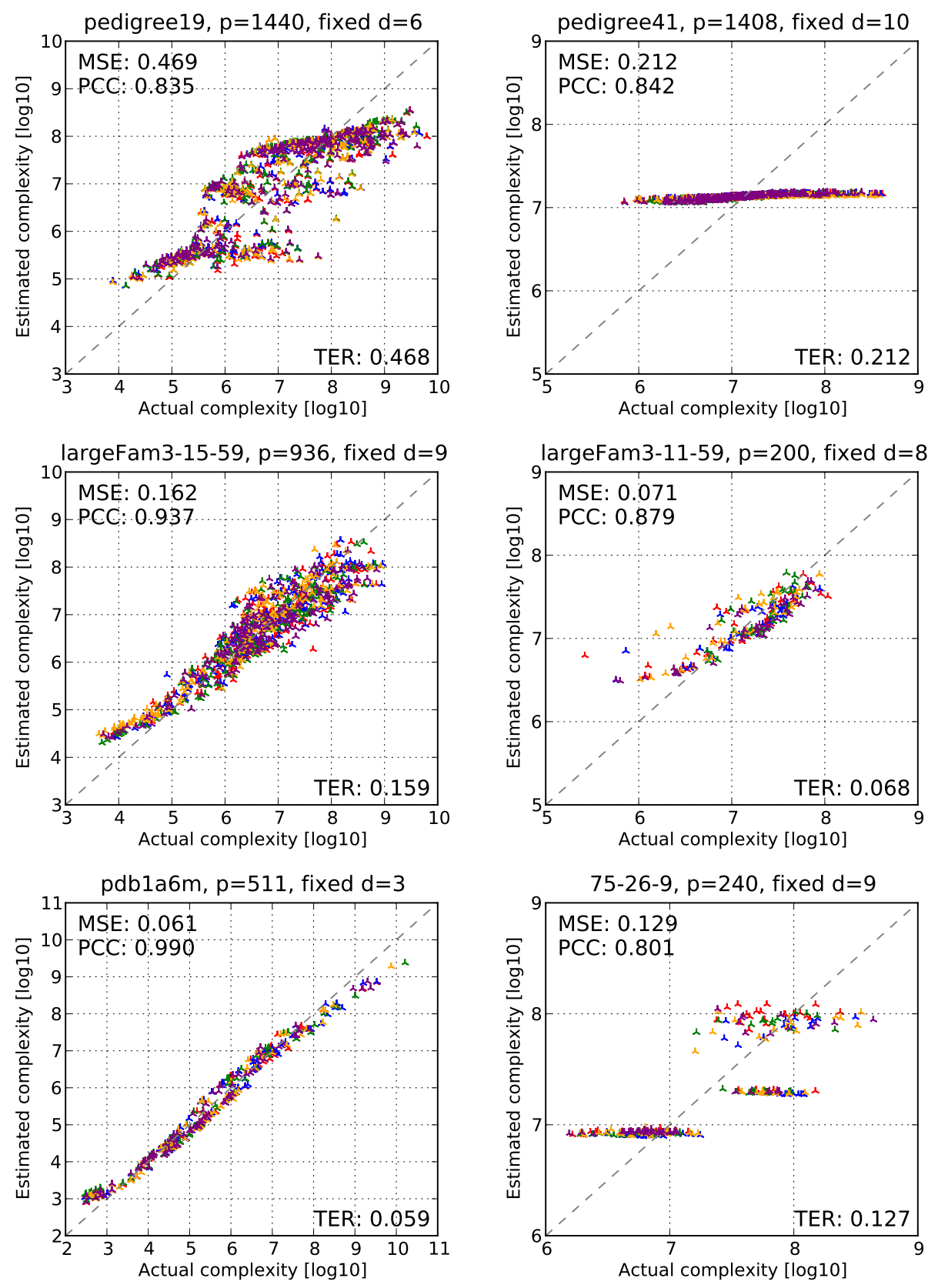

Figure 3: Actual vs. predicted subproblem complexity when learning per problem instance, using 5-fold cross validation.

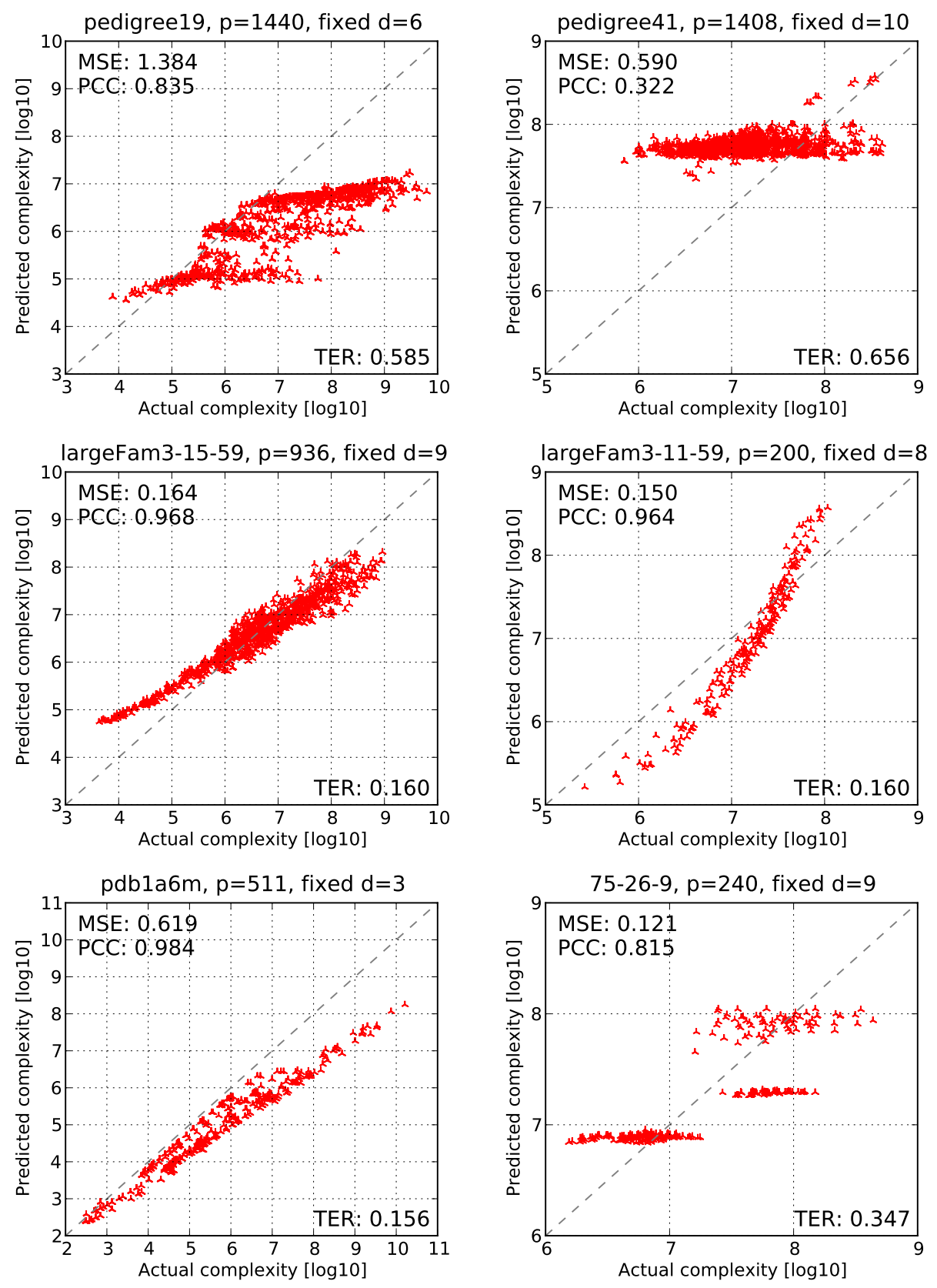

Figure 4: Actual vs. predicted subproblem complexity when learning per problem class.

### 4.2 LEARNING PER PROBLEM CLASS

Secondly, we aim to assess how well we can learn a model for an entire problem class. That is, we learn only once from sample subproblems of instances from the problem class in question. For testing we perform cross validation on the level of problem instances. Namely, we predict the subproblems of a given instance by fitting a model using subproblem samples of other instances from the same class – but not of the test instance itself.

Results are shown in Figure 4. Compared to Figure 3 in the previous section, estimates for the grid and largeFam instance are very similar and yield almost the same mean squared error. MSE increases for the pedigree and largeFam instances, but the PCC and overall shape of the predictions also remain similar, with the exception of *pedigree41* which sees both MSE and PCC deteriorate.

### 4.3 LEARNING MULTIPLE PROBLEM CLASSES

Lastly, we investigate how good a model we can learn from subproblems of instances across multiple problem classes. In particular, given a problem instance we learn a regression model on the subproblems of all other instances, regardless of their problem class.

Results are given in Figure 5, analogous to Figures 3 and 4. The two pedigree problems see an improved MSE and PCC (significantly for *pedigree41*), but the other instances suffer from a slightly larger prediction error. However, we again note that the overall shape of the plots remains roughly linear, which is also captured by high PCC values.

### 4.4 MOST INFORMATIVE FEATURES

Linear regression has the advantage that the resulting models can be straightforward to interpret. Namely, to assess the informativeness of feature $\phi_i$ we simply look at the absolute value of its coefficient $\lambda_i$ in the regression model. Assuming a normalized sample set, features with larger absolute values contribute more to the predictions and are thus intuitively more informative.

In addition, recall that the $L_1$-regularization in lasso regression implicitly performs feature selection by assigning $\lambda_i = 0$ for some $i$. In our case, training on the entire sample set (11,500 subproblem instances, regularization parameter through cross-validation) yielded non-zero $\lambda_i$ for nine features, as shown in in Table 3. In addition each feature's

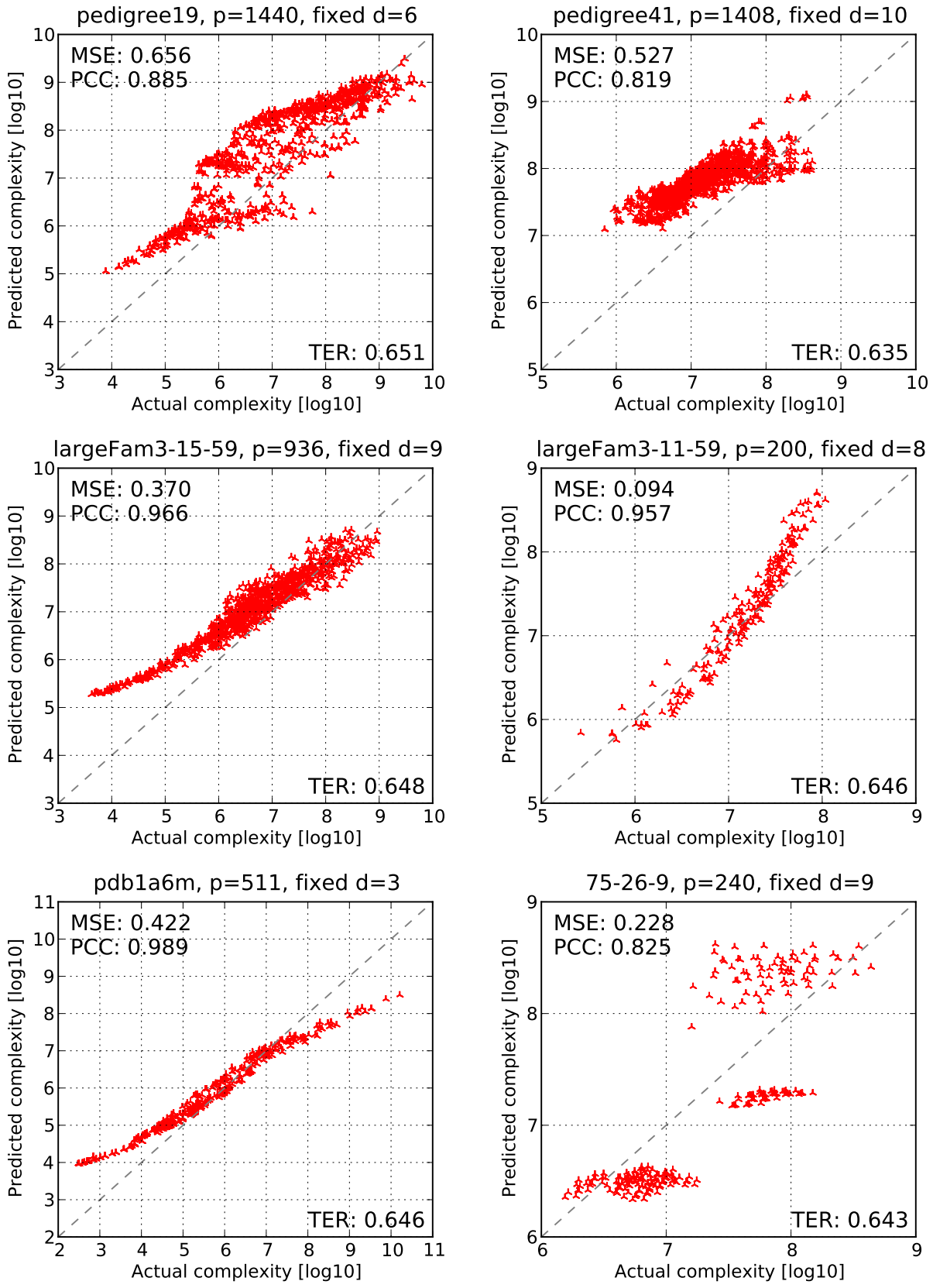

Figure 5: Actual vs. predicted subproblem complexity when learning across problem classes.

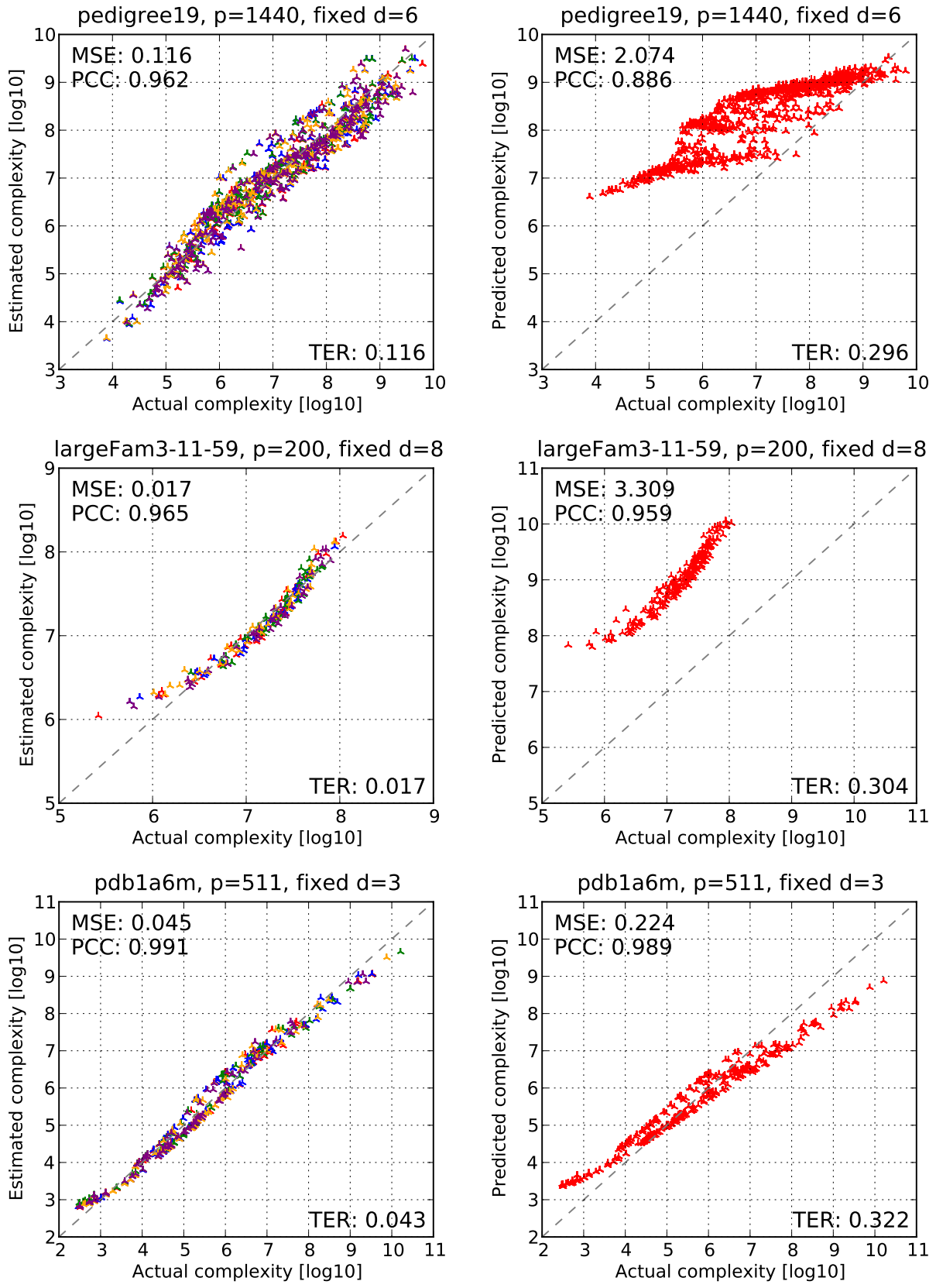

Figure 6: Example prediction results using a quadratic regression model. Left: learning per problem instance, using 5-fold cross validation (cf. Fig. 3). Right: learning from all problem classes (cf. Fig. 5).

*cost of omission* ("coo") as defined in [11] is given, which measures the normalized difference between the prediction error of the model with all nine features and the prediction error of a model trained with the respective feature omitted (using 5-fold cross-validation in all cases).

The particular set of features can be somewhat misleading, however, since lasso regression tends to pick only one of several highly correlated features. Yet it is useful to gain a conceptual understanding. In particular, we observe that the most informative features are dynamic, extracted from a limited AOBB probe or based on the initial subproblem bounds. Only the fifth feature, max. subproblem context size minus mini-bucket $i$-bound, is static, with a normalized cost of omission of only 16. This ties in to Section 2.2, where we observed that the asymptotic complexity bound of AOBB (based on static parameters) yields little information in this context.

| Feature $\phi_i$ | $\lvert\lambda_i\rvert$ | coo |
|---|---|---|
| Average branching degree in probe | 0.57 | 100 |
| Average leaf node depth in probe | 0.39 | 87 |
| Subproblem upper bound minus lower bound | 0.22 | 17 |
| Ratio of nodes pruned by heuristic in probe | 0.20 | 27 |
| Max. context size minus mini bucket $i$-bound | 0.19 | 16 |
| Ratio of leaf nodes in probe | 0.18 | 10 |
| Subproblem upper bound | 0.11 | 7 |
| Std. dev. of subproblem pseudo tree leaf depth | 0.06 | 2 |
| Depth of subproblem root node in overall space | 0.05 | 2 |

Table 3: Features present in the linear model trained by lasso regression, with their model coefficients $\lambda_i$ and their cost of omission "coo" (normalized).

### 4.5 NON-LINEAR REGRESSION

Here we briefly summarize results from our investigation of quadratic feature expansion, as detailed in Section 3.5. Selected prediction results for three instances are shown in Figure 6. On the left are results of learning per problem instances, on the right we plot the prediction accuracy when learning from all problem classes.

Comparing the plots on the left (per-problem learning) with Figure 3, we note that quadratic regression does a bit better than linear regression in terms of MSE and very similarly with regards to PCC. In contrast to [11], however, we find deteriorated prediction performance when comparing the plots on the right with Figure 5: while the PCC value is similar, the mean squared prediction error when learning from multiple problem classes increases considerably for *pedigree19* and *largeFam3-11-59*. Notably, however, the training error remains fairly low in both cases, which is indicative of overfitting. And indeed, with over 600 subproblem features and just 31 different instances, the quadratic regression model is likely to pick up specific characteristics of each instance that hurt its predictive performance.

Since quadratic models are also more expensive to train and lack the straightforward interpretability of a linear model, we feel that the latter is better-suited for our purposes.

### 4.6 INTERPRETATION OF RESULTS

We have trained and evaluated our proposed regression model on the three levels of learning laid out in Section 3.3, trading off between the wider applicability of the learned models and the challenges of capturing increasingly general sample sets. Learning per problem instance provided a good baseline but has limited relevance in practice, since each new instance requires extensive sampling of subproblems to train on. Learning per problem class is more reasonable as the learned model can be reused within the given problem class; our experiments showed good performance. Finally, learning across classes is the most challenging as the sample set is likely to be more diverse and have higher variance, requiring more training samples; however, once we learn a model it can be used throughout.

And indeed, our results in Section 4.3 show that, given our substantial set of 11,500 subproblem samples, the model can accommodate this most general level of learning across problem classes without a noticeable penalty, at least for the current collection of problem classes. A model learned across classes is therefore also the basis for the next section, where we demonstrate the benefit of robust complexity estimates in the context of distributed AOBB.

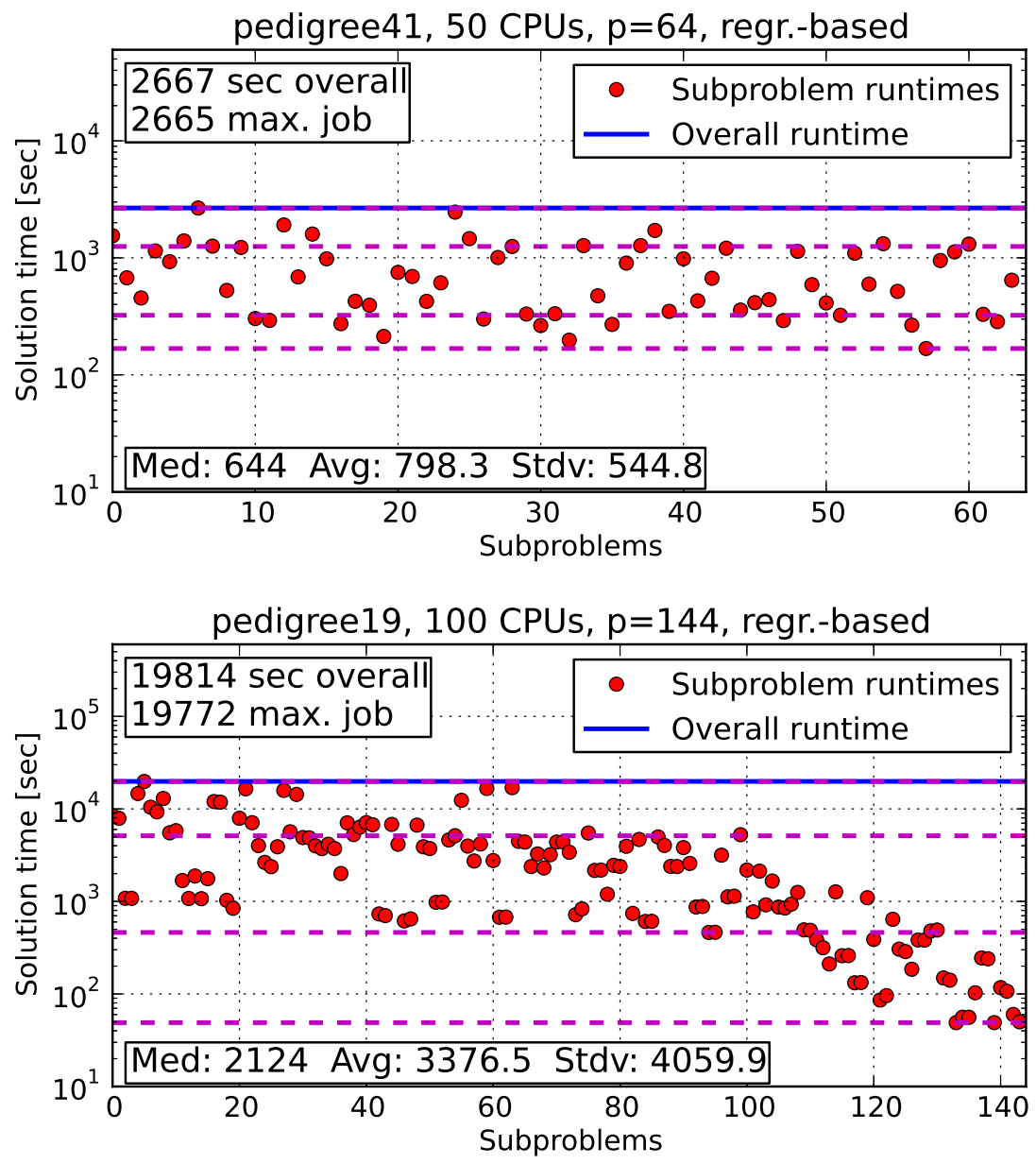

Figure 7: Subproblem statistics for regression-based parallelization (cf. fixed-depth parallelization in Fig. 2), $p$ denotes the number of subproblems.

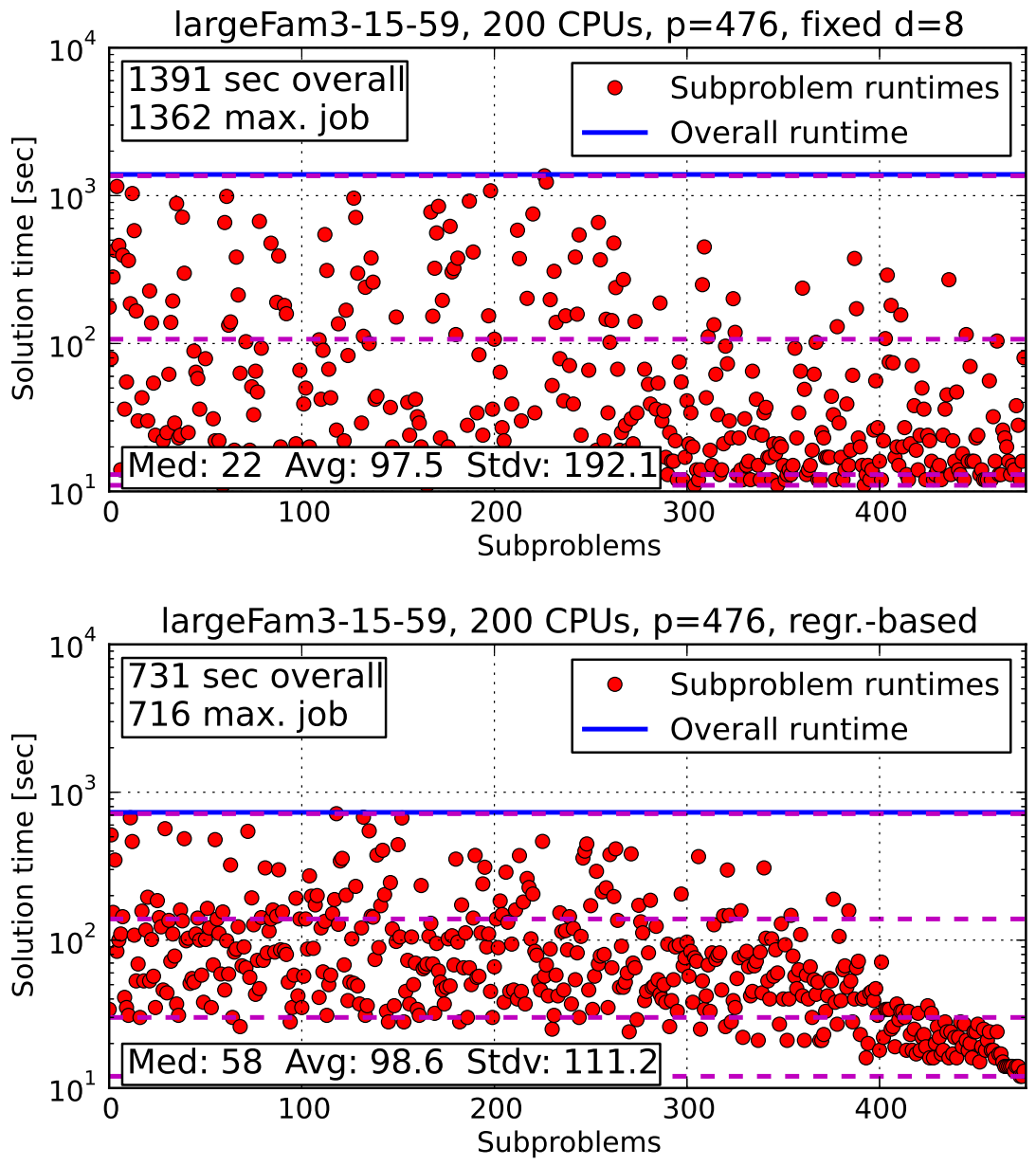

Figure 8: Subproblem statistics for fixed-depth (top) and regression-based (bottom) parallelization on *largeFam3-15-59* instance.

## 5 REGRESSION-BASED LOAD BALANCING IN PRACTICE

In this section we present selected experimental results that show the potential of the proposed regression models in guiding the parallelization process, as described in Section 2.2 – a comprehensive empirical evaluation of Distributed AOBB is beyond the scope of this paper. As noted above, the regression model used for experiments in this section was learned at the most general level, using all available problem classes as discussed in Section 4.3 (but always excluding the test problem instance).

### 5.1 IMPROVING LOAD BALANCING

To demonstrate the profound impact the complexity predictions can have on the load balancing of the parallel scheme, we revisit the two parallel experiments presented in Section 2.2, Figure 2. In both cases the overall performance was heavily dominated by very few long-running subproblems.

Figure 7 shows runtime statistics for parallel execution on these instances, using the regression model for load balancing. Figure 8 gives an additional example on a largeFam instance. In all cases we see that the max. subproblem runtime has been reduced greatly, close to 50% for *largeFam3-15-59* (1,362 to 716 seconds for “max. job” in Figure 8). We also note the drastically lower standard deviation in subproblem runtimes.

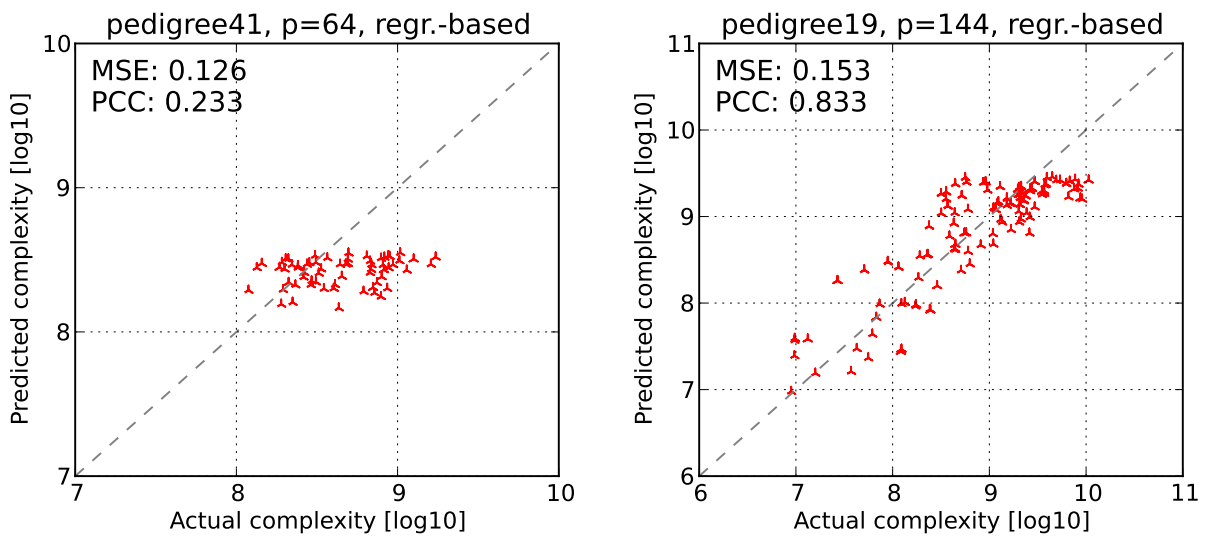

Figure 9: Actual vs. predicted subproblem complexity from the two parallel executions in Figure 7.

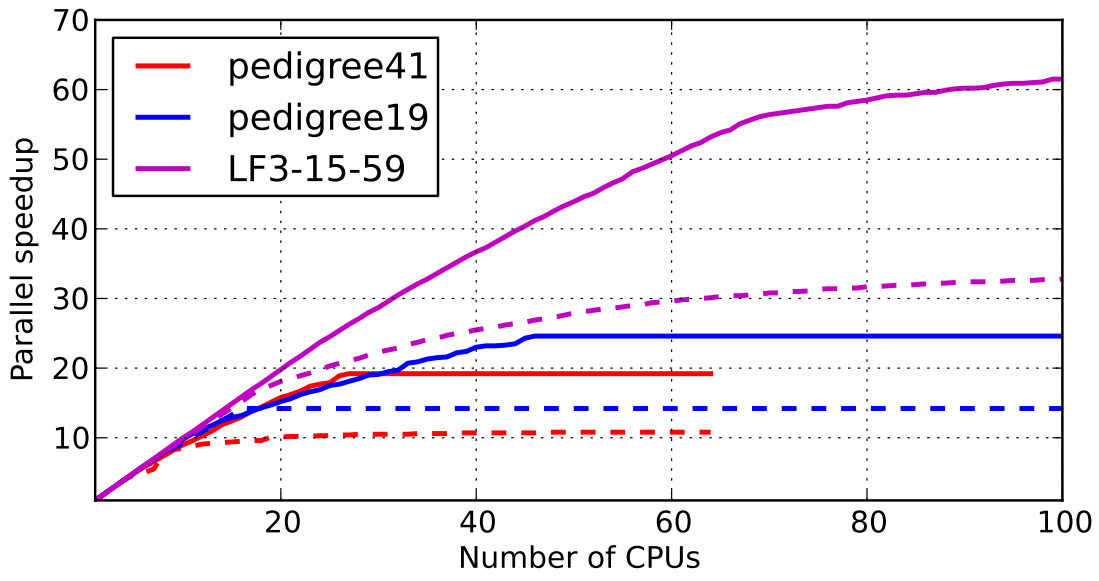

Figure 10: Parallel speedup for *pedigree41*, *pedigree19*, and *largeFam3-15-59* (cf. Figures 2, 7, and 8) as a function of the number of parallel CPUs. Dashed lines represent fixed-depth parallelization, solid lines correspond to parallel runs guided by our regression model.

In addition, Figure 9 compares the complexity estimates obtained during the parallel execution with the actual recorded values. In both cases we observe good prediction error and *pedigree19* in particular also shows good PCC.

### 5.2 FACILITATING PARALLEL SPEEDUP

Figure 10 plots the parallel speedup for the three problem instances considered in Section 5.1 over the number of parallel CPUs. For each instance we show (as a dashed line) the speedup when using fixed-depth parallelization and (with a solid line) the speedup of the parallel execution guided by the regression model.

Comparing each instance’s two entries reveals a clear advantage for the regression-based parallelization: it achieves higher speedups, roughly by a factor of two, and seems to plateau later, i.e. it is able to utilize a larger number of parallel resources.

On the other hand, most curves appear to level off well before their theoretical limit. This indicates that further improvements are possible, even though “perfect” speedup is unattainable in practice, since splitting a given subproblem often yields components of widely varying size and large jumps in complexity.

One way to mitigate these issues lies in increasing the subproblem granularity, i.e., setting the number of subproblems to match several times the number of parallel CPUs. However, this may add overhead in the general distributed context and redundancies in the particular graphical model context, which can negate potential gains in extreme cases. Indeed, finding the right balance in granularity is a central research issue in the field of distributed computing.

## 6 CONCLUSION & FUTURE WORK

We have presented a case study of complexity estimation in the context of parallelizing the state-of-the-art sequential optimization algorithm AND/OR Branch and Bound. The pruning power of the algorithm makes parallel load balancing very challenging, leading to inefficiencies in practice.

To address these symptoms we have proposed to employ statistical regression analysis in order to identify bottlenecks for parallel performance ahead of time. In particular, we developed a linear regression model that uses a variety of static as well as dynamic features to predict a subproblem’s complexity, enabling us to detect and split problematic subproblems.

We identified three distinct levels of learning and evaluated our proposed model accordingly, using more than 11,000 subproblem samples from 31 problem instances and four problem classes. Results were good throughout, with generally low prediction error and high correlation coefficients.

In the context of our parallel scheme, we have shown how the regression model can enable more effective load balancing and improved parallel speedup. This last set of results, however, also outlined opportunities for further improvements and future research, including varying the parallel granularity.

Future work with respect to learning of complexity estimates will expand to more instances and additional problem classes. In that context we also plan to investigate how our learned models perform on instances from previously unseen problem classes. Furthermore, we are trying to devise more subproblem features, for instance extracted directly from the cost function tables within a subproblem.

### ACKNOWLEDGEMENTS

Work supported in part by NSF grants IIS-0713118, IIS-1065618 and NIH grant 5R01HG004175-03.